\documentclass{article}
\usepackage{spconf,amsmath,graphicx,hyperref,multirow}
\usepackage{booktabs}
\usepackage{amssymb}

\usepackage{enumitem}

\title{\emph{SURE}: \underline{S}ynergistic \underline{U}ncertainty-aware \underline{RE}asoning for Multimodal Emotion Recognition in Conversations}

\name{Yiqiang Cai$^{1,2\ast}$\thanks{$\ast$ Equal contribution.}~
Chengyan Wu$^{1,2\ast}$~
Bolei Ma$^{3\dagger}$~
Bo Chen$^{4}$~
Yun Xue$^{1,2\dagger}$\thanks{$\dagger$ Corresponding authors: bolei.ma@lmu.de, xueyun@m.scnu.edu.cn}~
Julia Hirschberg$^{5}$~
Ziwei Gong$^{5}$}
  
  \address{\ninept$^{1}$Guangdong Provincial Key Laboratory of Quantum Engineering and Quantum Materials \\
      \ninept$^{2}$School of Electronic Science and Engineering (School of Microelectronics), 
South China Normal University \\
      \ninept$^{3}$LMU Munich \& Munich Center for Machine Learning ~
      $^{4}$Shenzhen University ~
      $^{5}$Columbia University}
      
\begin{document}
\ninept
\maketitle
\begin{abstract}
Multimodal emotion recognition in conversations (MERC) requires integrating multimodal signals while being robust to noise and modeling contextual reasoning. Existing approaches often emphasize fusion but overlook uncertainty in noisy features and fine-grained reasoning. We propose \textbf{SURE} (\textbf{S}ynergistic \textbf{U}ncertainty-aware \textbf{RE}asoning) for MERC, a framework that improves robustness and contextual modeling. SURE consists of three components: an \emph{Uncertainty-Aware Mixture-of-Experts} module to handle modality-specific noise, an \emph{Iterative Reasoning} module for multi-turn reasoning over context, and a \emph{Transformer Gate} module to capture intra- and inter-modal interactions. Experiments on benchmark MERC datasets show that SURE consistently outperforms state-of-the-art methods, demonstrating its effectiveness in robust multimodal reasoning. These results highlight the importance of uncertainty modeling and iterative reasoning in advancing emotion recognition in conversational settings. \footnote{Code is available at: \url{https://github.com/swaggy66/SURE}.}
\end{abstract}

\section{Introduction and Background}
\label{sec:intro}

Multimodal Emotion Recognition in Conversations (MERC) \cite{wu2025multi, wu-etal-2025-beyond} focuses on identifying emotions in dialogue utterances by integrating textual, acoustic, and visual signals. This task is crucial for applications such as social media understanding \cite{DBLP:conf/ic3/KumarDD15, gong25b_interspeech}, healthcare support \cite{DBLP:conf/riiforum/PujolMM19, liu-etal-2025-propainsight}, and empathetic human-computer interaction \cite{DBLP:journals/coling/ZhouGLS20,DBLP:journals/taffco/HuHHX23,10.1145/3613904.3642336}. Unlike conventional emotion recognition on isolated utterances, MERC requires modeling both contextual dependencies across turns and speaker-specific dynamics, making it more challenging yet also more informative.

Deep learning has driven substantial progress in MERC, with recent methods focusing on richer multimodal fusion and contextual reasoning. For example, SDT \cite{DBLP:journals/tmm/MaWLZZX24} employs intra- and inter-modal Transformers with gated fusion, while DF-ERC \cite{DBLP:conf/mm/Li0LZTCJL23} integrates features via dual-level decoupling and contextual re-fusion. MultiEMO \cite{DBLP:conf/acl/ShiH23} enhances recognition of minority emotions with cross-attention and specialized loss design, and UniMSE \cite{DBLP:conf/emnlp/HuLZLWL22} exploits contrastive learning to model coherence across modalities. Other works, such as JOYFUL \cite{DBLP:journals/corr/abs-2311-11009}, CMCF-SRNet \cite{DBLP:conf/acl/ZhangL23}, MM-NodeFormer \cite{DBLP:conf/interspeech/HuangML24}, and MMPCGN \cite{DBLP:journals/ijon/MengSADLL24}, further explore multimodal correlations via diverse transformer or graph-based mechanisms. Collectively, these approaches highlight the importance of advanced fusion strategies and contextual modeling in MERC.

Despite these advances, two fundamental challenges remain. First, real-world conversational data inevitably contains noise, yet most methods overlook modality-specific and sample-level uncertainty. Second, contextual and emotional cues are often retrieved in a static manner, which hinders fine-grained reasoning and disrupts logical consistency across dialogue turns. Addressing these challenges is essential for improving robustness and accuracy in MERC, particularly in dynamic, real-world conversational settings..

To this end, we propose \textbf{SURE} (\textbf{S}ynergistic \textbf{U}ncertainty-aware \textbf{RE}asoning), a framework that explicitly incorporates uncertainty modeling and iterative reasoning into multimodal dialogue understanding. SURE consists of three key components: (1) an \emph{Uncertainty-Aware Mixture-of-Experts} module that dynamically mitigates modality-specific noise, (2) an \emph{Iterative Reasoning} module that performs multi-turn reasoning over context, and (3) a \emph{Transformer Gate} module that adaptively captures intra-modal and inter-modal interactions. Together, we find that they produce robust multimodal representations that better capture emotional dynamics in conversations.

Our main contributions are threefold:
\begin{itemize}[leftmargin=*]
    \item We propose a novel framework SURE and demonstrate its consistent outperformance over state-of-the-art methods on benchmark MERC datasets.
    \item SURE introduces three synergistic modules: an uncertainty-aware mixture-of-experts (MoE) module for robust noise mitigation, an iterative reasoning module inspired by cognitive processes, a transformer-based multimodal gate for adaptive fusion.
    \item Beyond performance gains, we provide in-depth analysis and interpretability studies, demonstrating how SURE enables more transparent and controllable emotional reasoning across modalities and context turns.

\end{itemize}

\begin{figure*}[t] 
\centering 
\includegraphics[width=0.765\textwidth]{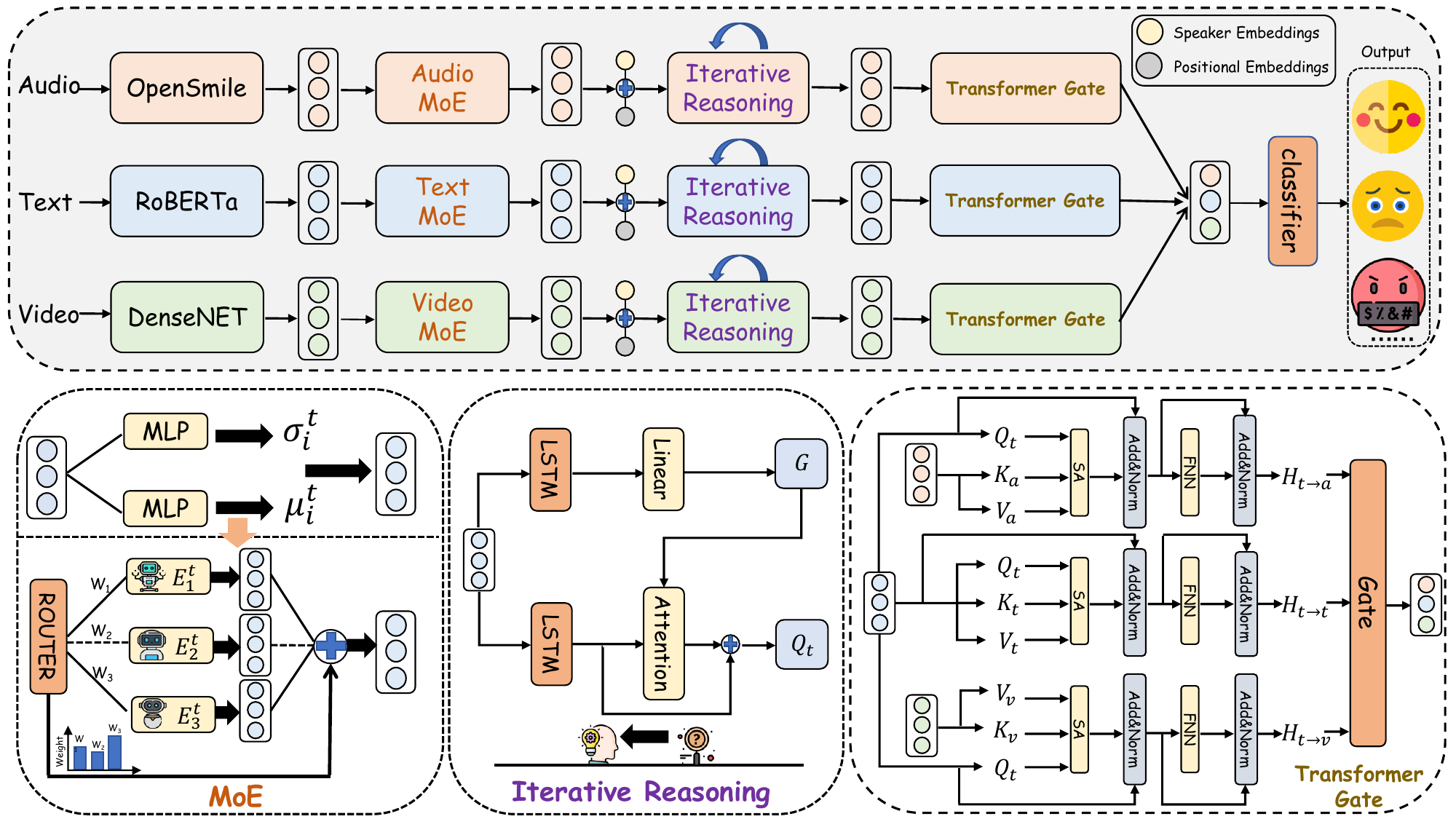} 
\caption{The SURE Framework. The top shows the overall framework of multimodal inputs and classifying process. The bottom consists of three main components: Uncertainty-Aware MoE, Iterative Reasoning, and Transformer Gate, corresponding to the three major intermediate modules in the SURE framework.} 
\label{fig:SURE Framework} 
\end{figure*}

\section{Methodology - The \emph{SURE} Framework}
\label{sec:method}

\newenvironment{shrinkeq}[2]
{ \bgroup
  \addtolength\abovedisplayshortskip{#1}
  \addtolength\abovedisplayskip{#1}
  \addtolength\belowdisplayshortskip{#2}
  \addtolength\belowdisplayskip{#2}}
{\egroup\ignorespacesafterend}

\subsection{Problem and Framework}

\textbf{Problem Formulation.} Given a dialogue composed of N consecutive utterances $D = \{u_1, u_2, \dots, u_N\}$ with $M$ speakers, the MERC task aims to assign each utterance $u_i$ an emotion label $e_i \in \mathcal{Y}$ from a predefined category set. Each $u_i$ contains textual ($u_i^t \in \mathbb{R}^{d_t}$), acoustic ($u_i^a \in \mathbb{R}^{d_a}$), and visual ($u_i^v \in \mathbb{R}^{d_v}$) features. The goal is to leverage these multi-modal signals and conversational context to predict $e_i$.

\noindent\textbf{Overall Framework.}
The overall architecture of the proposed SURE Framework for MERC is illustrated in Figure \ref{fig:SURE Framework}. After
extracting utterance-level unimodal features, the framework consists of three core modules: an uncertainty-aware MoE module (§\ref{sec:moe}), an Iterative Reasoning module (§\ref{sec:icm}), and a Transformer Gate module (§\ref{sec:MultiTrans}), followed by an emotion classifier (§\ref{sec:classifier}).

\subsection{Utterance-Level Feature Extraction }
\noindent\textbf{Textual Modality.}
We employ the RoBERTa Large model to extract textual features. RoBERTa is a pre-trained model based on a multi-layer Transformer encoder architecture, developed from BERT, which can efficiently capture textual representations. We fine-tune RoBERTa on conversational transcript data for the emotion recognition task and use the embeddings of the [CLS] token from the last layer as the textual features.

\noindent\textbf{Acoustic Modality.}
We adopt openSMILE for acoustic feature extraction. openSMILE is a versatile signal processing toolkit that offers a scriptable console application, enabling flexible configuration and integration of modular feature extraction components.

\noindent\textbf{Visual Modality.}
We employ DenseNet, pre-trained on the Facial Expression Recognition Plus dataset, for visual feature extraction. DenseNet is a highly effective convolutional neural network (CNN) architecture composed of multiple densely connected blocks, each consisting of several layers.


\subsection{Uncertainty-Aware MoE}
\label{sec:moe}
In the MERC task, considering the impact of noisy data on modal features, we follow \cite{GaoH00SX24} to enhance the experts’ ability to capture uncertainty when processing samples with varying levels of noise. The experts are primarily composed of parallel sub-networks (typically MLPs) and are activated via a routing mechanism across different feature spaces or tasks.

Specifically, after feature extraction, we obtain multimodal features $X_i^m, m \in \{v, a, t\}$, where $i$ denotes the sample instance. We map the features of each modality to a diagonal multivariate Gaussian. Specifically, we define the representation $z_i^m$ of each sample $x_i^m$ in the latent space as a Gaussian distribution, expressed as:
\begin{equation}
p(z_i^m \mid x_i^m) \sim \mathcal{N}(\mu_i^m,\, \sigma_i^{m2} I)
\end{equation}

where the mean and variance are predicted via two independent fully connected layers:

\begin{equation}
\mu_i^m = f_{\theta_1^m}, \quad \sigma_i^m = f_{\theta_2^m}
\end{equation}

Each feature can then be sampled as:
\begin{equation}
z_i^m = \mu_i^m + \epsilon \sigma_i^m, \quad \epsilon \sim \mathcal{N}(0, I)
\end{equation}
where $\mu_i^m$ represents the stable feature representation, and $\sigma_i^m$ quantifies the stochastic uncertainty of the modality feature.

We can now quantify the uncertainty in noisy data based on the magnitude of $\sigma_i^m$, and subsequently integrate it into MoE. The MoE consists of two primary components: the experts ($E$) and the gating network ($G$). The experts process the input samples effectively, while the gating network determines the most appropriate experts to activate for processing samples with specific levels of noise. 

Given a feature $x_i^m$ as input, the gating network selects the top $k$ performing experts and retains their corresponding outputs $E$ for subsequent processing. The routing algorithm of the gate can be expressed as:
\begin{equation}
G(x_i^m) = \text{TOP}_k \big( \text{softmax}(\text{Linear}(x_i^m)) \big)
\end{equation}
where the output dimension of the Linear layer equals the number of experts. The $\text{TOP}_k$ operation sets all but the top $k$ values to zero. We select the top $k$ experts based on the highest softmax scores, and use these scores as the weights for the experts’ outputs. When the total number of experts is $N$, the process can be formalized as:
\begin{equation}
\small
z_i^m = \sum_{j=1}^{N} G_j^m(x_i^m) E_j^m(x_i^m)
\end{equation}

So far, we have obtained a dynamic network capable of capturing uncertainty. Next, we enable the gating network to select experts based on uncertainty. Specifically, we use the magnitude of $\sigma_i^{m2}$ to quantify the uncertainty in noisy data. During forward propagation, each expert generates a $\sigma_i^m$ to represent its uncertainty.The gate ultimately selects experts with lower uncertainty to process the corresponding data.
   

\subsection{Iterative Reasoning}
\label{sec:icm}

Single-step feature modeling remains insufficient to capture the complex emotional dynamics in dialogues.
Inspired by emotion cognition theories \cite{Schachter1962,Scherer2001}, we design an Iterative Reasoning Module following the MoE to extract emotional cues from dialogue context. High-quality modality features $z_i$ are encoded via an LSTM and a linear layer to form a global memory $G = \{g_i\}_{i=1}^N$:
\begin{equation}
g_i = \text{Linear}(\text{LSTM}(z_i, h_{i-1}))
\end{equation}
where $z_i$ denotes the modality feature of the $i$-th sample, taking any modality as an example, $h_{i-1}$ is the previous hidden state.
\begin{equation}
r_i^{(t)} = \text{Attention}(q_i^{(t)}, G)
\end{equation}
\begin{equation}
q_i^{(t+1)} = [q_i^{(t)}; r_i^{(t)}]
\end{equation}
A query vector $q$ retrieves relevant contextual cues from $G$ using an attention mechanism: where $r_i^{(t)}$ is the contextual information retrieved at iteration $t$.
The initial query is derived from the utterance-level context $z_i$:
\begin{equation}
q_i^{(0)} = W_q z_i + b_q, \quad h_i^{(0)} = 0
\end{equation}
where $W_q$ and $b_q$ are learnable parameters, and $h_i^{(0)}$ is the initial hidden state of the iterative reasoning LSTM.

Finally, a second LSTM performs iterative reasoning, repeatedly updating the query vector to produce refined emotional cues:
\begin{equation}
q_i^{(t+2)}, h_i^{(t+2)} = \text{LSTM}(q_i^{(t+1)}, h_i^{(t)})
\end{equation}
where $q_i^{(t+1)}$ is the query vector at iteration $t+1$, and $h_i^{(t)}$ is the corresponding hidden state. After multiple iterations, the final output query vector $U_i$ serves as the refined emotional cue for the $i$-th utterance in the dialogue.

\subsection{Transformer Gate}
\label{sec:MultiTrans}
Emotional expression typically relies on multimodal collaboration, while unimodal reasoning alone is insufficient to effectively capture intra- and inter-modal dependencies within the utterance sequence. 
To this end, we propose the Transformer Gate Module, which consists of two levels:intra-modal attention and inter-modal attention.We employ self-attention (SA) to model intra-modal dependencies and cross-attention (CA) to capture inter-modal interactions. Both built upon the standard Transformer encoder~\cite{vaswani2017attention}.

Taking the reasoned text modality $U_{t}$ as an example, in the intra-modal attention, $U_{t}$ are used as queries, keys, and values to capture intra-modal dependencies:
\begin{equation}
\tilde{U}_{t \to t} = \text{SA}(U_{t}, U_{t}, U_{t})
\end{equation}
\begin{equation}
\bar{U}_{t \to t} = \text{Norm}(\tilde{U}_{t \to t} + U_t)
\end{equation}
\begin{equation}
U_{t \to t} = \text{Norm}(\text{FFN}(\bar{U}_{t \to t}) + \bar{U}_{t \to t})
\end{equation}
In the inter-modal attention, the text modality serves as the query, while the acoustic ($a$) and visual ($v$) modalities act as keys and values to integrate complementary information and model inter-modal dependencies:
\begin{equation}
\tilde{U}_{m \to t} = \text{CA}(U_{t}, U_{m}, U_{m}), \quad m \in \{a,v\}
\end{equation}
\begin{equation}
\bar{U}_{m \to t} = \text{Norm}(\tilde{U}_{m \to t} + U_t)
\end{equation}
\begin{equation}
U_{m \to t} = \text{Norm}(\text{FFN}(\bar{U}_{m \to t}) + \bar{U}_{m \to t})
\end{equation}
Finally, the model adaptively fuses the representations enhanced by intra- and inter-modal attention through a gating mechanism to obtain the final text representation:
\begin{equation}
H_{t} = Gate(U_{t \to t}, U_{a \to t}, U_{v \to t})
\end{equation}
Similarly, the final representations for the acoustic and visual modalities, $H_a$ and $H_v$ are obtained.
\subsection{Emotion Classifier}
\label{sec:classifier}

To predict the emotion of each utterance, we first concatenate the textual representation $H_t$, acoustic representation $H_a$, and visual representation $H_v$ to obtain a fused vector:
\begin{equation}
H = [H_t ; H_a ; H_v]
\end{equation}
where $[\,;\,]$ denotes the concatenation operation.  

The fused vector $H$ is then fed into a fully connected layer followed by a softmax layer to obtain the emotion probability vector for each utterance:
\begin{equation}
E = W_e H + b_e, \quad \hat{y}_i = \text{softmax}(E)
\end{equation}
where $W_e$ and $b_e$ are learnable parameters. The predicted emotion label is obtained by:
\begin{equation}
\hat{e}_i = \arg\max(\hat{y}_i).
\end{equation}


\section{Experiments and Results}

\subsection{Experimental Settings}

\noindent\textbf{Datasets.}
We validate the performance of SURE on two benchmark datasets: IEMOCAP \cite{DBLP:journals/lre/BussoBLKMKCLN08} and MELD \cite{DBLP:conf/acl/PoriaHMNCM19}. The statistics of the datasets are shown in Table \ref{tab:datasets}.
\begin{table}[htbp]
\scriptsize
\setlength\tabcolsep{9pt}
\caption{Datasets statistics.}
\centering
\begin{tabular}{lcccc}
\hline
& \multicolumn{2}{c}{\textbf{IEMOCAP (6-Ways)}} & \multicolumn{2}{c}{\textbf{MELD (7-Ways)}} \\
\cmidrule(lr){2-3} \cmidrule(lr){4-5}
\textbf{Datasets} & \#Dialogue & \#Utterance & \#Dialogue & \#Utterance \\
\hline
Train+Val & 120 & 5810 & 1153 & 11098\\
Test & 31 & 1623 & 280 & 2610\\
\hline
\end{tabular}
\label{tab:datasets}
\end{table}

\noindent\textbf{Evaluation Metrics.} Following prior studies \cite{DBLP:conf/mm/Li0LZTCJL23,DBLP:conf/acl/ShiH23}, we use overall accuracy and weighted average F1-score, and additionally report those on each emotion category for a more detailed evaluation.

\noindent\textbf{Baselines.}
We conduct a comprehensive comparison between SURE and a set of competitive baseline methods \cite{wu2025multi}. For fundamental graph-based methods, we select MMGCN \cite{DBLP:conf/acl/HuLZJ20}, MM-DFN \cite{DBLP:conf/icassp/HuHWJM22}, GS-MCC \cite{Meng2024RevisitingME}, Joyful \cite{DBLP:journals/corr/abs-2311-11009} and MMPCGN \cite{DBLP:journals/ijon/MengSADLL24}; for fusion-based approaches, we select SDT \cite{DBLP:journals/tmm/MaWLZZX24}, DialogueTRM \cite{DBLP:conf/emnlp/Mao0WGL21},
DF-ERC \cite{DBLP:conf/mm/Li0LZTCJL23} and MM-NodeFormer \cite{DBLP:conf/interspeech/HuangML24} as state-of-the-art baselines.

\noindent\textbf{Implementation Details.}
We use PyTorch framework with AdamW optimizer on two RTX A6000 GPUs. 
For training, the batch size is set to 16 for IEMOCAP and 32 for MELD, with 150 and 50 training epochs, respectively. 
The learning rate is configured as $1\times10^{-4}$ for IEMOCAP and $5\times10^{-6}$ for MELD, while a dropout rate of $0.5$ is applied across all experiments. 
Moreover, in the MoE module, we employ a Top-$k$ routing strategy with $k=3$. The reported results of our approach are the average of 10 runs.


\subsection{Results}

Table~\ref{tab:results} summarizes the experimental results on the two datasets, from which we draw the following observations. 
Our comparison focuses on parameter-efficient, discriminative MERC models and does not include MLLM-based approaches due to their fundamentally different modeling paradigm.

    (1) \textbf{SURE consistently outperforms all baselines on both IEMOCAP and MELD. } Notably, the performance gain on IEMOCAP is more substantial, likely due to the increased speaker variability and contextual complexity present in MELD \cite{Zha2024EsihgnnEI}. These results highlight SURE’s robust modeling of emotional dynamics through uncertainty-aware fusion and context reasoning, enabling better generalization across diverse datasets.
    
    
    
    (2) \textbf{Within graph-based methods, SURE achieves significant improvements over all baselines.} Specifically, on IEMOCAP, it surpasses Joyful by 4.76\% and 3.77\% in ACC and F1, respectively; on MELD, it outperforms MERC-GCN by 4.82\% in F1. These findings underscore the effectiveness of SURE’s structural modeling, which better captures the relational dynamics of dialogue compared to existing graph-based baselines.
    
      (3) \textbf{Compared to existing fusion-based methods, SURE also consistently outperforms the baselines.} On MELD, it improves over SDT by 0.42\% in ACC and 0.76\% in F1, suggesting that existing fusion-based approaches often overlook the perception and handling of modality noise. In contrast, our proposed method dynamically selects the expert with the lowest uncertainty, thereby extracting more reliable features. On IEMOCAP, SURE achieves 1.07\% and 0.6\% gains in ACC and F1 over MM-NodeFormer, which primarily relies on the text modality. The performance gain can be attributed to the context reasoning module’s multi-round inference mechanism, which enhances the retrieval of emotional cues and enables more nuanced emotion reasoning, outperforming simpler gating-based context aggregation methods.
      
In summary, SURE uniquely integrates multimodal fusion with uncertainty-aware expert selection and multi-round context reasoning. This allows it to extract more reliable features, navigate modality noise, and capture fine-grained emotional cues, and achieve consistent gains across both graph- and fusion-based baselines.

\begin{table}[tbp]
\footnotesize
\centering
\caption{Comparison with baselines on IEMOCAP and MELD.}
\label{tab:results}
\begin{tabular}{lccccc}
\hline
\multirow{2}{*}{\textbf{Models}}  
& \multicolumn{2}{c}{\textbf{IEMOCAP}} & \multicolumn{2}{c}{\textbf{MELD}} \\
\cline{2-5}
 &  Acc & F1 & Acc & F1 \\
\hline
 &   & Graph-based Methods &  &  \\
 \hline
MMGCN       &   66.36   & 66.26 &   60.42   & 58.31 \\
MM-DFN        &   68.21  &  68.18 &  62.49   & 59.46 \\
Joyful     &   70.55&   71.03  &  62.53   & 61.77 \\
MMPCGN     &  68.90  &  68.00 &  60.70  & 59.30\\
MERC-GCN          &   --   & 68.98 &   --  & 62.54 
\\
\hline
 &   & Fusion-based Methods &  &  \\
 \hline
DialogueTRM        & 69.50  & 69.70 & 65.70  & 63.50 \\
DF-ERC    & 71.84  & 71.75 & \textbf{68.28}  & 67.03 \\
SDT        & 73.95 & 74.08 & 67.55 & 66.60 \\
MM-NodeFormer   & 74.24 & 74.20 &  67.86 & 66.09  \\
\hline
\textbf{SURE} (ours)   & \textbf{75.31} & \textbf{74.80} & 67.97 & \textbf{67.36} \\
\hline
\end{tabular}
\end{table}


\subsection{Ablation Study}
In order to evaluate the performance of the model under various modalities for both datasets and to assess the contribution of the key modules to the overall performance, a detailed ablation study is performed, as shown in Table \ref{tab:ablation study}.

\textbf{Effects of the MoE and Cognitive Reasoning Modules.}
When the MoE module is removed, the performance of SURE drops considerably, with F1 scores decreasing by approximately 0.57\% and 0.34\% on the two datasets. This highlights the crucial role of dynamically modeling noisy data for robust feature representation, particularly in capturing modality-specific uncertainties to mitigate the influence of cross-modal noise. Moreover, removing the iterative reasoning module leads to an additional performance decline of about 0.38\% and 0.44\%, respectively. These results indicate that consciously and sequentially reasoning over contextual information enables the effective integration of emotional cues, thereby improving the overall performance.

\textbf{Effects of Missing Modality Combinations on Model Robustness.}
We further investigate the model performance under different missing-modality combinations across various noisy scenarios. The results demonstrate that when text modality is present, the model consistently achieves better performance (e.g., the text-only setting surpasses the audio-only and vision-only settings by 10.65\% and 28.89\% in F1 on IEMOCAP, respectively). In contrast, the inclusion of the visual modality tends to introduce larger performance fluctuations. This can be attributed to the fact that textual information provides a more direct and stable reflection of the speaker’s emotions and intentions, whereas visual signals are more susceptible to noise. Moreover, the combinations involving text with either audio or vision consistently outperform the combination of audio and vision alone (e.g., with text+audio achieves a 11.70\% higher F1 than vision+audio on IEMOCAP). The complete modality combination yields the best performance, confirming the effectiveness of SURE.
\begin{table}[!tbp]
\caption{Model variations and fusion network ablation results.}
\footnotesize
\renewcommand\arraystretch{1.0}
\centering
\begin{tabular}{lcccc}
\hline
& \multicolumn{2}{c}{\textbf{IEMOCAP}} & \multicolumn{2}{c}{\textbf{MELD}} \\
\cmidrule(lr){2-3} \cmidrule(lr){4-5}
\textbf{Set-ups} & Acc & F1 & Acc & F1 \\
\hline
\multicolumn{5}{l}{\textit{\textbf{Methods}}} \\
SURE & 75.31 & 74.80 & 67.97 & 67.36 \\
~w/o MoE & 74.99 & 74.23 & 67.65 & 67.02 \\
~w/o Reasoning & 75.02 & 74.42 & 67.32 & 66.92 \\
\hline
\multicolumn{5}{l}{\textit{\textbf{Modality}}} \\
Text &68.66  &68.39  &66.16  &66.29  \\
Audio &60.13  &57.74  &37.21  &39.88  \\
Visual&42.32  &39.50  &30.86  &31.34  \\
Text + Audio &73.98  &73.05  &66.37  &66.44  \\
Text + Visual &69.42  &68.89  &65.94  &66.15  \\
Visual + Audio &62.20  &61.35  &38.36  &40.54  \\
\hline
\end{tabular}
\label{tab:ablation study}
\end{table}








\section{Conclusion}
We propose \textbf{SURE}, a novel framework for multimodal emotion recognition in conversations that combines uncertainty-aware Mixture-of-Experts with enhanced global reasoning. By dynamically routing unimodal features through specialized experts, SURE effectively mitigates modality-specific noise, while the reasoning module captures fine-grained cross-modal interactions. Extensive experiments on benchmark datasets demonstrate that SURE outperforms state-of-the-art baselines.  In future work, we plan to explore more adaptive expert-sharing strategies and extend the framework to broader multimodal dialogue understanding tasks.

\section{Acknowledgment}
This work was partially supported by the National Natural Science Foundation of China (No. 62573298), the Guangdong Provincial Key Laboratory (No. 2023B1212060076), the Shenzhen Key Laboratory of Media Security (No. SYSPG20241211174032004), the Guangdong Basic and Applied Basic Research Foundation under Grant 2023A1515011370, the National Natural Science Foundation of China (32371114), the Characteristic Innovation Projects of Guangdong Colleges and Universities (No. 2018KTSCX049), and the National Science Foundation under ARNI 
(The NSF AI Institute for Artificial and Natural Intelligence) via PHY-2229929. 


\bibliographystyle{IEEEbib}

\bibliography{refs}

\end{document}